\documentclass[sigconf]{acmart}
\usepackage{markdown}
\usepackage{listings}
\usepackage{xcolor}

\lstdefinelanguage{Markdown}{
  morekeywords={},
  sensitive=false,
  morecomment=[l]{\#},
  morestring=[b]"
}

\AtBeginDocument{%
  }

\setcopyright{acmlicensed}
\copyrightyear{2025}
\acmYear{2025}
\acmDOI{XXXXXXX.XXXXXXX}
\acmConference[FDG '26]{Foundations of Digital Games 2026}{August 10--13, 2026}{Copenhagen}
\acmISBN{978-1-4503-XXXX-X/2018/06}

\begin{document}

\title{Vox Deorum: A Hybrid LLM Architecture for 4X / Grand Strategy Game AI - Lessons from Civilization V}

\author{John Chen}
\email{johnchen@arizona.edu}
\affiliation{%
  \institution{University of Arizona}
  \country{United States}
}

\author{Sihan Cheng}
\email{sihancheng2026@u.northwestern.edu}
\affiliation{%
  \institution{Northwestern University}
  \country{United States}
}

\author{Can Gurkan}
\email{gurkan@u.northwestern.edu}
\affiliation{%
  \institution{Northwestern University}
  \country{United States}
}

\author{Ryan Lay}
\email{rlay2@wisc.edu}
\affiliation{%
  \institution{University of Wisconsin-Madison}
  \country{United States}
}

\author{Moez Salahuddin}
\email{moezsalahuddin221@gmail.com}
\affiliation{%
  \institution{Independent Researcher}
  \country{United States}
}

\renewcommand{\shortauthors}{anonymized et al.}

\begin{abstract}
Large Language Models' capacity to reason in natural language makes them uniquely promising for 4X and grand strategy games, enabling more natural human-AI gameplay interactions such as collaboration and negotiation. However, these games present unique challenges due to their complexity and long-horizon nature, while latency and cost factors may hinder LLMs' real-world deployment. Working on a classic 4X strategy game, Sid Meier's Civilization V with the Vox Populi mod, we introduce Vox Deorum, a hybrid LLM+X architecture. Our layered technical design empowers LLMs to handle macro-strategic reasoning, delegating tactical execution to subsystems (e.g., algorithmic AI or reinforcement learning AI in the future). We validate our approach through 2,327 complete games, comparing two open-source LLMs with a simple prompt against Vox Populi's enhanced AI. Results show that LLMs achieve competitive end-to-end gameplay while exhibiting play styles that diverge substantially from algorithmic AI and from each other. Our work establishes a viable architecture for integrating LLMs in commercial 4X games, opening new opportunities for game design and agentic AI research.
\end{abstract}


\begin{teaserfigure}
  \includegraphics[width=\textwidth]{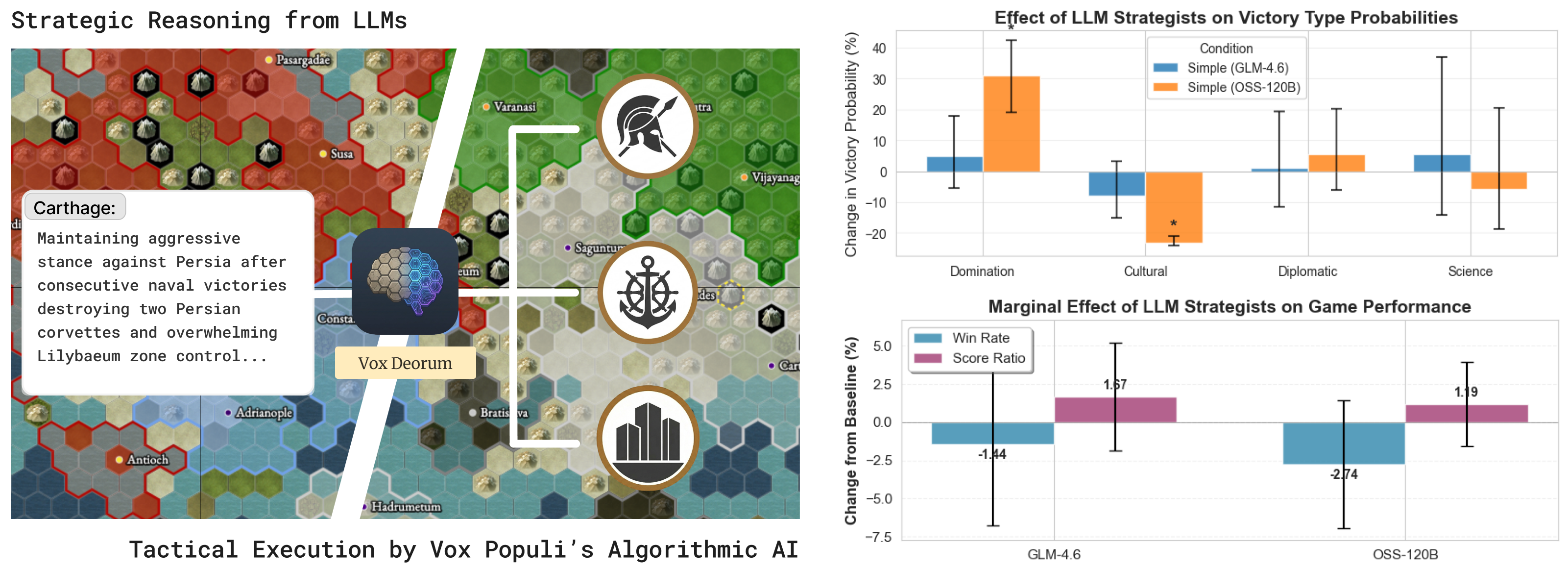}
  \caption{Vox Deorum is a hybrid LLM+X architecture for 4X strategy games. In the left panel, we see how the LLM strategist processes game state and sets high-level strategies that guide Civilization V (with Vox Populi)'s algorithmic AI for tactical execution. With a simple prompt, LLM strategists achieved a comparable win rate and score ratio against the baseline AI (top-right) through distinct victory types (bottom-right) and playstyles (see findings).}
  \label{fig:teaser}
\end{teaserfigure}


\maketitle

\section{Introduction}
4X and grand strategy games are among the most complex environments for human players. Named after their core mechanics (eXploration, eXpansion, eXploitation, and eXtermination), these games challenge players to manage political entities ranging from clans to civilizations across hundreds of turns. Under conditions of imperfect information and multilateral competition, players have to make strategic decisions across multiple aspects each turn: economic development, military strategy, diplomatic relations, and technological advancement. With such depth and emergent complexity, popular game titles such as Sid Meier's Civilization, Europa Universalis, and Stellaris have captivated millions of players. 

Yet, AI opponents in those games struggle to match human expectations, often relying on major handicaps (i.e., bonuses only available to AI players) to stay competitive (e.g., \cite{civ5_difficulty}). Effective 4X gameplay requires two distinct competencies: high-level strategic reasoning and tactical execution. Algorithmic AI may excel at tactical execution, as evidenced by the Vox Populi project's improvement on Civilization V's AI \cite{vice_civ5_ai_modders}. However, 4X games present a uniquely difficult challenge at the strategic level, with multi-agent interactions, multiple victory paths, imperfect information, and general-sum game dynamics. As such, algorithmic AI systems struggle to achieve the flexibility and unpredictability that human opponents naturally provide, creating long-standing challenges for game designers and developers. 

Advanced AI approaches, such as large language models (LLMs) and reinforcement learning (RL), offer promising avenues for addressing these limitations; however, each approach faces distinct challenges when applied in isolation. RL agents have achieved superhuman performance in games like Go \cite{silver2016go} and StarCraft II \cite{Vinyals2019starcraft}. However, a previous study finds RL models underperforming in 4X games, inclining towards short-term rewards while failing to plan for long-term gains \cite{civrealm}. On the other hand, LLMs bring complementary strengths: they can reason and take instructions in natural languages, articulate multi-step plans, adapt to novel situations, and often provide out-of-the-box usability \cite{hu2025llmgamesurvey, sweetser2024llmgames}. However, while LLMs can perform reasonably well in mini-game tasks \cite{civrealm,wang2025digitalplayer} and outperform RL agents \cite{civrealm}, both approaches fail to survive full games. The volume of tactical decisions required each turn (managing dozens of cities and units) also creates prohibitive costs and latency for real-world deployment.

To combine the strengths of LLMs and other AI systems, this paper introduces Vox Deorum, a hybrid LLM+X architecture that delegates tactical execution to specialized subsystems (``X'') while reserving strategic decision-making for LLMs. We validate our architecture through an empirical study on Sid Meier's Civilization V with the Vox Populi community mod, which has substantially enhanced the game's built-in AI \cite{vice_civ5_ai_modders}. We evaluated two open-source models' out-of-the-box gameplay capability with a simple prompt design, establishing a baseline for further research. Specifically, we asked the following research questions:

\begin{itemize}
    \item \textbf{RQ1:} Can hybrid LLM+X architectures handle end-to-end long-horizon gameplays in commercial 4X games?
    \item \textbf{RQ2:} How do open-source LLMs perform vs. Vox Populi's algorithmic AI?
    \item \textbf{RQ3:} How do LLMs play differently compared to each other or Vox Populi's algorithmic AI?
\end{itemize}

Our findings demonstrate the viability and potential of the proposed LLM+X architecture. Both of the open-source LLMs that were considered (GPT-OSS-120B from OpenAI and GLM-4.6 from z.ai) successfully completed hundreds of full games, with survival rates approaching 100\%. Across 2,327 full games, LLMs achieved statistically tied win rates and score ratios against Vox Populi's algorithmic AI (VPAI) baseline. Both LLMs exhibited distinct play styles compared with VPAI, with different victory type preferences, strategic pivoting behaviors, and policy adoption trajectories. With excellent latency and reasonable cost, this architecture opens a wide range of opportunities for game design, (machine learning) research, and commercial adoption. Our study makes the following contributions:

\begin{itemize}
    \item We introduce Vox Deorum, a hybrid LLM+X architecture for 4X games, delegating tactical execution to complementary modules for better latency, cost-effectiveness, and gameplay performance.
    \item We present an open-source implementation of the Vox Deorum architecture for Civilization V with Vox Populi, providing a research platform for game AI and agentic AI researchers.
    \item We conduct the largest empirical study to date of LLM agents in 4X games, comparing end-to-end gameplay across 2,327 complete games between LLMs and the baseline VPAI.
\end{itemize}

\section{Related Work}
Recent research on advanced game AI often focuses on reinforcement learning (RL)-based or large language models (LLM)-based agents, two prominent approaches that offer complementary strengths in tactical execution and high-level strategic reasoning \cite{silver2016go, silver2018alphazero, Vinyals2019starcraft, berner2019dota2, sweetser2024llmgames, hu2025llmgamesurvey, zhang2024llm, duan2024gtbenchuncoveringstrategicreasoning, costarelli2024gamebenchevaluatingstrategicreasoning, akata25, yannakakis2025aigames}.

\subsection{RL-based agents for Strategy Games}
RL-based agents have achieved superhuman performance in chess and RTS games. Self-play deep RL and search agents attain beyond human play in board games such as Go, chess, and shogi, establishing a performance ceiling for tactical decision making \cite{silver2016go, silver2017mastering, silver2018alphazero}. In high-dimensional, partially observable real-time environments, agents such as AlphaStar and OpenAI Five reach Grandmaster or world champion level in StarCraft II and Dota 2 \cite{Vinyals2019starcraft, berner2019dota2}. AlphaStar, for example, demonstrates that RL agents can control an enormous state and action space under imperfect information while coordinating many units \cite{Vinyals2019starcraft}. Once trained, these policies are fast at inference and directly optimized for objectives such as win rate or cumulative score, which makes them strong tactical executors within their training distributions \cite{silver2018alphazero, Vinyals2019starcraft, berner2019dota2}.

However, RL's adoption in long-horizon, multilateral strategy games (e.g., Civilization) remains challenging. Training of RL models requires massive computation and game-specific engineering, yet the resulting policies are often tied to a single game ruleset \cite{silver2018alphazero, Vinyals2019starcraft, berner2019dota2}. RL agents often overfit to particular training environments and struggle to transfer to new levels or altered configurations \cite{cobbe2020procgen}, limiting their performance in novel games or dynamic strategic situations. While RL agents perform well in fixed-team, zero-sum games, they are less examined in 4X and grand strategy games, where teams are emergent and fluid, cooperation and competition often co-exist, and human-facing negotiation or coordination is essential. In particular, inspecting, interpreting, or steering black-box RL policies can be difficult \cite{puiutta2020explainablereinforcementlearningsurvey, rudin2019stop}, complicating RL's adoption in Civilization-like games. 

\subsection{LLM-based agents for Strategy Games}
While early exploration of LLM-based agents has uncovered promising high-level reasoning potential in strategy games, scaling these systems to complex, long-horizon environments can be challenging. Hu et al. \cite{hu2025llmgamesurvey} and Sweetser et al. \cite{sweetser2024llmgames} catalog LLM-based game agents across components such as perception, memory, planning, role-playing, action execution, and learning. Their reviews span a broad set of genres that includes action \cite{xu2025agentsplaythousands3d}, adventure \cite{ngaw2024openended, phan2025textquestsgoodllmstextbased}, role-playing \cite{vanStegeren2021gpt2quests, shao-etal-2023-character}, simulation \cite{park2023generativeagentsinteractivesimulacra}, and strategy games \cite{LLM_spire, LLM_pysc2}. Most applications use LLMs for NPC dialogue \cite{vanStegeren2021gpt2quests, treanor2025sliceoflife}, question answering \cite{zhu2023calypso, kelly2023ttrpg}, content generation \cite{freiknecht2020stories, todd2023levelgen, alnassar2023questville, nasir2024word2worldgeneratingstoriesworlds}, or lightweight decision making \cite{sidji2024codenamesassistant, chi2024amongagentsevaluatinglargelanguage}. More recently, Google's SIMA 2 \cite{sima2} extends this line of work by demonstrating a generalist, multimodal agent capable of operating across a wide range of environments using natural language instructions, visual perception, and action grounding. However, despite these advances, such systems remain at an exploratory stage, with only a few studies having examined 4X or grand strategy games\cite{civrealm, wang2025digitalplayer}.

Pure LLM agents can articulate multi-step plans in games, yet they remain expensive and cost-ineffective for tactical decision-making. While a study on Slay the Spire shows that zero-shot LLMs can play complete runs and reason several turns ahead \cite{LLM_spire}, LLMs' tactical decisions remain weaker than a conventional search-based agent specialized for the task and are highly sensitive to prompt design. Poglitsch et al. demonstrate that LLM agents can participate in multi-round social deduction in Werewolf, with planning improving anticipatory reasoning and conversational coherence. However, complex prompts can degrade coherence, and model quality strongly influences error rates \cite{LLM_werewolf}. In real-time strategy settings, Li et al. introduce LLM-PySC2, which exposes StarCraft II through text and multimodal interfaces. LLMs can propose plausible high-level strategies and occasionally succeed in easier full-game scenarios, yet they struggle with micro-level control, often producing hallucinated or invalid actions, mis-targeted units, or unstable multi-agent communication \cite{LLM_pysc2}. Moreover, as each decision may require several seconds of LLM reasoning, latency can be a bottleneck for fine-grained control in such games \cite{LLM_pysc2}.

More recently, hybrid architectures that pair LLMs with complementary modules have gained traction among game AI researchers, which offer a more reliable path toward strategic competence. In Guandan, a four-player cooperative card game under imperfect information, Yim et al. find that augmenting LLMs with Theory-of-Mind prompts and an RL-based action recommender substantially improves cooperation and win rate. With these scaffolds, the GPT-4 agent attains a level of play comparable to the state-of-the-art RL model DanZero+ \cite{yim2024guandan}, while weaker LLMs remain below the RL baseline. Such results suggest that while LLMs are well-suited for high-level reasoning and semantic abstraction, they require complementary mechanisms for gameplay. To operate robustly in complex, long-horizon games, a hybrid LLM+X architecture is necessary to overcome the limitations of pure LLMs. 

\subsection{Advanced AI for 4X Games, e.g. Civilization}
Although recent studies have brought LLMs to multiple genres of games, only a few studies have explored their potential in 4X or grand strategy games, such as Civilization. Our preliminary review identified two papers so far: CivRealm (on FreeCiv, a Civilization II remake) \cite{civrealm}, and Digital Player (on Unciv, a Civilization V remake) \cite{wang2025digitalplayer}. With a pure LLM or RL approach, both papers advance LLMs' capabilities in micro-level tasks, yet both fall short of end-to-end gameplay. The CivRealm study identifies 4X games' ever-increasing complexity as the main challenge \cite{civrealm, wang2025digitalplayer}. As the game progresses, its state keeps growing in complexity. The map gradually reveals; the number of cities and units grows; new technologies and policies unlock more buildings and diplomatic options. With imperfect information for each player, the combinatorial space of possible decisions grows exponentially. Each choice has both short-term tactical and long-term strategic consequences, while full games can last many hours to days for human players. Moreover, such games are often multilateral without fixed teams (i.e., friends and enemies are both temporary and can shift under circumstances), making them uniquely challenging to work on. 

The CivRealm study presents a technical system and empirical study to incorporate pure LLM or RL agents into FreeCiv \cite{civrealm}. The system exposes tensor-based APIs for RL agents and language-based APIs for LLM agents. It provides a benchmark environment with a suite of mini-games for development, battle, and diplomacy, as well as a full-game mode. Pure RL agents achieve reasonable performance in the mini-games, yet they tend to adopt short-sighted strategies in full games (e.g., over-producing units to boost short-term score while sacrificing long-term development and expansion). CivRealm’s LLM agents demonstrate relatively better high-level behaviors, yet struggle with essential gameplay tasks such as defense, economic balance, and crisis handling. Both paradigms struggle to survive (and therefore complete) the full game. Based on the findings, the authors highlight the potential of hybrid architectures, which can combine the strengths of language-based reasoning and tactical control (through RL or other pathways) \cite{civrealm}.

More recently, Digital Player has attempted to bring LLM-based workflows into Unciv, incorporating tool usage, retrieval-augmented generation (RAG), and self-reflection mechanisms \cite{wang2025digitalplayer}. In particular, the study highlighted the impact of in-game simulators (that predict the outcome of a given action), showing the hybrid architecture's potential for augmenting LLMs' strategic decision-making capabilities. However, the study adopts a simplified ruleset (e.g., only conquering victory is possible, and many limitations in the rules are lifted), and LLMs can only take in-game diplomatic actions. Without a comparison between LLM-based agents and either a human or an AI baseline, it is still unclear whether those workflows may succeed in end-to-end Civilization games. 

\section{System Design}
This section presents Vox Deorum's hybrid LLM+X architecture and our sample implementation. By focusing LLMs on macro-level decision-making and delegating tactical execution to ``X'' (complementary modules, e.g., algorithmic AI or RL-based AI), we believe this novel architecture can balance performance, latency, and cost in the context of complex strategic games while opening future interactive design opportunities.

\subsection{Hybrid LLM+X Architecture}
Our novel LLM+X architecture integrates two elements to efficiently embed Generative AI into 4X or grand strategy games: an LLM element that handles high-level strategic planning and an ``X'' element that handles tactical execution. In Vox Deorum, the LLM is responsible for reviewing overall situations and conducting high-level decision-making for playing the Civilization V game. The LLM sets the grand strategy (the victory type that the AI player targets), economic and military strategies (which unit, building to prioritize, and the aggressiveness of military actions), next technology to research, next policy to adopt, as well as diplomatic personas (the tendency of the AI player to declare war, make friendship, behave deceptively, and so on). Then, Vox Populi's algorithmic AI (VPAI) manages tactical executions, such as unit movement, combat targeting, city production queues, and tile improvements. \href{https://github.com/LoneGazebo/Community-Patch-DLL}{Vox Populi} (aka Community Patch Project) is a popular Civilization V mod which greatly expands the base game's strategic depth (e.g., the introduction of diplomatic vassals and more recent inclusion of 3/4th unique per-civilization components) and tactical AI. It also provides a well-maintained codebase for LLM integration. That said, the ``X'' element is not limited to algorithmic AI: our architecture can also accommodate rule-based AI, RL-based AI agents, and more. 

We believe our design has distinct advantages over prior pure-LLM or non-LLM approaches, which our paper will empirically validate:

\begin{itemize}
    \item \textbf{Performance and steerability}: Prior approaches with LLM or RL alone struggle to survive in Civilization games \cite{civrealm, wang2025digitalplayer}. In particular, even powerful LLMs still struggle with tactical decision-making where other AI forms excel \cite{kolasani2025llm}. By delegating tactical execution to VPAI, our design can improve LLM performance while keeping design opportunities upon LLMs' superior steerability and natural language capabilities.
    \item \textbf{Cost-effectiveness and latency}: Suppose the LLM needs to operate on each micro-level decision; the cost and latency for each turn can easily skyrocket. A single player in a Civilization V game can easily have $\sim$10 cities and $\sim$50 units mid-game, creating pressure on generation throughput. By focusing LLMs on high-level decision-making on a per-turn basis, our design can simultaneously increase cost-effectiveness and improve latency, opening pathways for real-world adoption.
\end{itemize}

\begin{figure}[t]
  \centering
  \includegraphics[width=\linewidth]{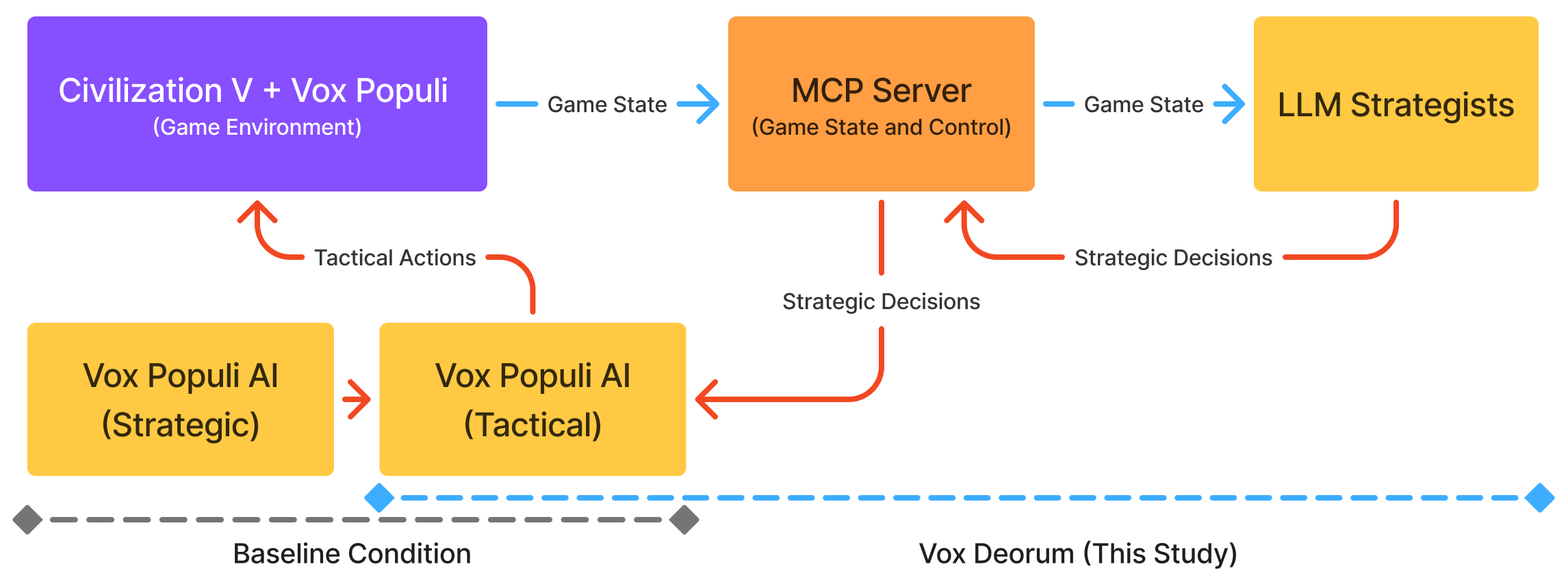}
  \caption{An overview of the Vox Deorum system, as implemented in this study.}
  \label{fig:architecture}
\end{figure}

\subsection{Vox Deorum System}
Instead of requiring the LLM to go through each micro-level decision and replacing the entire VPAI, our architecture only swaps the VPAI's strategic module with an LLM. Fig. \ref{fig:architecture} provides an overview of Vox Deorum's technical implementation, which is built upon the open-source mod Vox Populi (Community Patch) of Civilization V. To summarize, we exposed the game's internal mechanism from a Windows Named Pipe into a REST API. Then, we created a downstream MCP server to expose high-level functionalities (e.g., retrieving military states or overriding VPAI's strategies). Finally, we created an MCP client that activates an LLM strategist. The LLM strategist activates after each player's turn concludes, when it reviews the game state and sets high-level strategies for the VPAI to follow during the upcoming turn. We intentionally designed the timing to minimize inference latency: while the LLM processes, other players' turns proceed normally. A human player can take several minutes to play a turn, leaving abundant time for the LLM to complete its reasoning.

Encoding Civilization's rich game state for LLM input and output requires careful representation choices. Prior work has mainly explored and recommended text-based game representations for LLMs. For example, Bateni and Whitehead \cite{LLM_spire} represent Slay the Spire using card descriptions and key numeric statistics, finding that concise state summaries outperform verbose encodings. Poglitsch et al.'s Werewolf framework \cite{LLM_werewolf} and Yim et al.'s Guandan agent \cite{yim2024guandan} feed dialogue logs or pruned move lists as textual prompts. On the other hand, LLM-PySC2 \cite{LLM_pysc2} exposes StarCraft II through both text and multimodal observations. However, the paper reports hallucinated and invalid actions alongside unstable multi-agent control. Building on those studies, Vox Deorum's design encodes each turn's game state into a structured Markdown document with the following sections: Victory Progress; Strategic Options and Current Choices; Player Summaries; City Summaries; Military Summaries; and Events (since the last decision-making turn). Combined with other optimizations, the Markdown format reduces input token usage by up to two-thirds. The LLM has the same level of information visibility as any human or VPAI players.

Once the LLM strategist takes the input, it is tasked with choosing from a range of strategic options, policies, or technologies, also provided in the prompt. To maintain parity for our controlled experiment, this study's implementation reused VPAI's existing macro-level strategies. For example, the ``LosingMoney'' economic strategy lowers VPAI's numerical tendency to expand armies, while increasing the chance for building money-making buildings. Wherever LLMs are making decisions, VPAI's corresponding strategic modules are disabled. While we only experimented with a very simple prompt to understand LLMs' out-of-the-box capabilities, our abstraction of LLM strategists allows for much more complicated workflows, such as memory management and multi-agent collaboration.

The same tactical modules of VPAI then execute LLM strategists' macro-level decisions, as if VPAI's own strategic modules made them. Those modules are mostly implemented as search-based algorithms, covering micro-level decision-making processes in unit deployment, city management, diplomatic actions, etc. Except for policies and technologies, which LLMs directly choose from a few available options, LLMs' strategic decisions are mostly reflected in the ``flavor'' numbers - i.e., weight modifiers in the search algorithms. For example, ``WinningWars'' military strategy would make VPAI's tactical algorithm more likely to launch offensive actions, while ``EmpireDefenseCritical'' strategy would make it focus on homeland defense. 

Vox Deorum is \href{https://anonymous.4open.science/r/vox-deorum-fdg/README.md}{an open-source project}. intended for distribution and adoption among Civilization V players and Game AI researchers.

\section{Methodology}
\subsection{Environment Setup}
We conducted experiments using Civilization V and the Vox Populi mod (VP, version 5.0.0-alpha3), a community-developed overhaul that significantly expands game rules and improves AI performance. All games start on the most popular community map for VP, the Communitas\_79a, at a tiny size with four players. VPAI or LLMs controlled the macro-level strategies of Player 0, which in turn directed VPAI's tactical decision-making. Standard VPAI consistently controlled Players 1-3. All victory conditions were enabled: Domination, Cultural, Diplomatic, Science, and Time. In each game, 43 Civilizations were randomly assigned to players.

To evaluate LLMs' out-of-the-box strategic reasoning capabilities and provide a baseline for future research, we intentionally designed a minimal prompt-based approach for this study (see Appendix~\ref{app_sys_prompt} for the full prompt). LLMs can only see the current state of the game, including events from the previous turn. They can also see their last strategic decision and rationale, which serves as a very short-term memory.

\subsection{Evaluation Framework}
Our study addresses the following research questions:

\begin{itemize}
    \item \textbf{RQ1:} Can hybrid LLM+X architectures handle end-to-end long-horizon gameplays in commercial 4X games?
    \item \textbf{RQ2:} How do open-source LLMs perform vs. Vox Populi's algorithmic AI?
    \item \textbf{RQ3:} How do LLMs play differently compared to each other or Vox Populi's algorithmic AI?
\end{itemize}

We compared three conditions across 2,327 complete Civilization V games: VPAI, serving as the baseline (919 games); GPT-OSS-120B (983 games; 117 billion parameters; 5.1 billion activated; hosted by Jetstream2); and GLM-4.6 (425 games; 355 billion parameters; 32 billion activated; hosted by Chutes.ai). Games with irrecoverable crashes (mostly due to VP mod's alpha status) or excessive inference API endpoint failures (e.g., server unavailable that leads to a strategy gap of $15+$ turns) were excluded from analysis. We analyzed:

\begin{itemize}
    \item \textbf{Gameplay performance} through win rate, which captured whether Player 0 achieved any victory condition; and score ratio, which measured Player 0's highest score relative to that of all players. Score is calculated by Civilization V itself, combining multiple factors ranging from population, cities, wonders, military, policy, and technology.
    \item \textbf{Strategic behavior} through victory type distributions (which victory conditions were achieved); grand strategy adoption (proportion of game time spent targeting each victory type); strategy and persona change rates (frequency of strategic pivots); and policy trajectories (sequences of in-game policy branch adoptions).
    \item \textbf{Cost and latency} through input and output token statistics.\footnote{Due to a technical issue, this statistic is missing from $\sim$30\% of games.} Since latency depends on the actual hardware hosting the models and can fluctuate with workload levels, we derived it from token usage and existing hardware benchmarks instead of reporting the actual data.
\end{itemize}

While civilizations were randomly assigned across all conditions, we conducted fixed-effects regression analyses to further control for Civilization V's civilization-dependent effects. We used deviation (sum) coding, which centers estimates relative to the mean civilization effect. For binary outcomes (win rate, victory type achieved, policy adoption), we used logistic regression with $L1$ regularization to reduce collinearity. For continuous outcomes (score ratio, strategy adoption ratios), we used ordinary least squares regression. Token usage growth patterns were analyzed through polynomial regression to characterize scaling behavior across game progression. Confidence intervals are reported at 95\% throughout.

\section{Results}
This section presents aggregated results from our study. Appendix~\ref{app_prompts} provides a sample interaction with LLMs. Appendix~\ref{app_outputs} showcases a succinct sample playthrough rendered by Vox Deorum's companion project, Vox Deorum Replay Player.

\subsection{RQ1: Hybrid Architecture Viability}
Except for the rare irrecoverable crashes from Vox Populi itself, LLMs completed all games with a similar survival rate compared with the baseline. Despite differences between models, token usage and latency are both within a reasonable range. Yet, as games progress into later stages, input token usage grows quadratically (as opposed to output token usage, which grows linearly), placing pressure on context windows.

Our hybrid LLM+X architecture completes all games with a survival rate statistically tied with VPAI. On average, players controlled by OSS-120b (97.5\%) and GLM-4.6 (97.6\%) both survived as long as VPAI (97.3\%). Average game lengths were similarly stable across conditions (378.2, 377.4, and 371.8 turns).

Token usage differs between the two models. On average, OSS-120B consumes 20.35 million input tokens and 555,267 output tokens per game, while GLM-4.6 consumes 11.68 million input tokens and 247,881 output tokens. Since OSS-120B does not have built-in support for simultaneous tool calling, it needs more cycles to change multiple strategic parameters (e.g., strategy and policy) per turn. Using OpenRouter's pricing, an average game with OSS-120B would cost $\sim$\$0.86 per game (assuming $\$0.04/$mil for input and $\$0.2/$mil for output, as of 12/2025), which would cover many hours of human play time. Average token usage per turn (and thus, expected latency) remains reasonable. For example, based on Perplexity's early benchmark on OSS-120B (\url{https://www.perplexity.ai/hub/blog/gpt-oss-on-day-0}), an average turn with the model will cost 14.8 seconds: 1.98 seconds to prefill 52,854 tokens, and 12.8 seconds to generate 1,481 tokens. Such a speed is well within the range of Civilization V's built-in multiplayer timer, which gives each player 20 seconds of base time (in addition to more time per city and unit) under its fastest "blazing" setting.

\begin{figure}[]
  \centering
  \includegraphics[width=\linewidth]{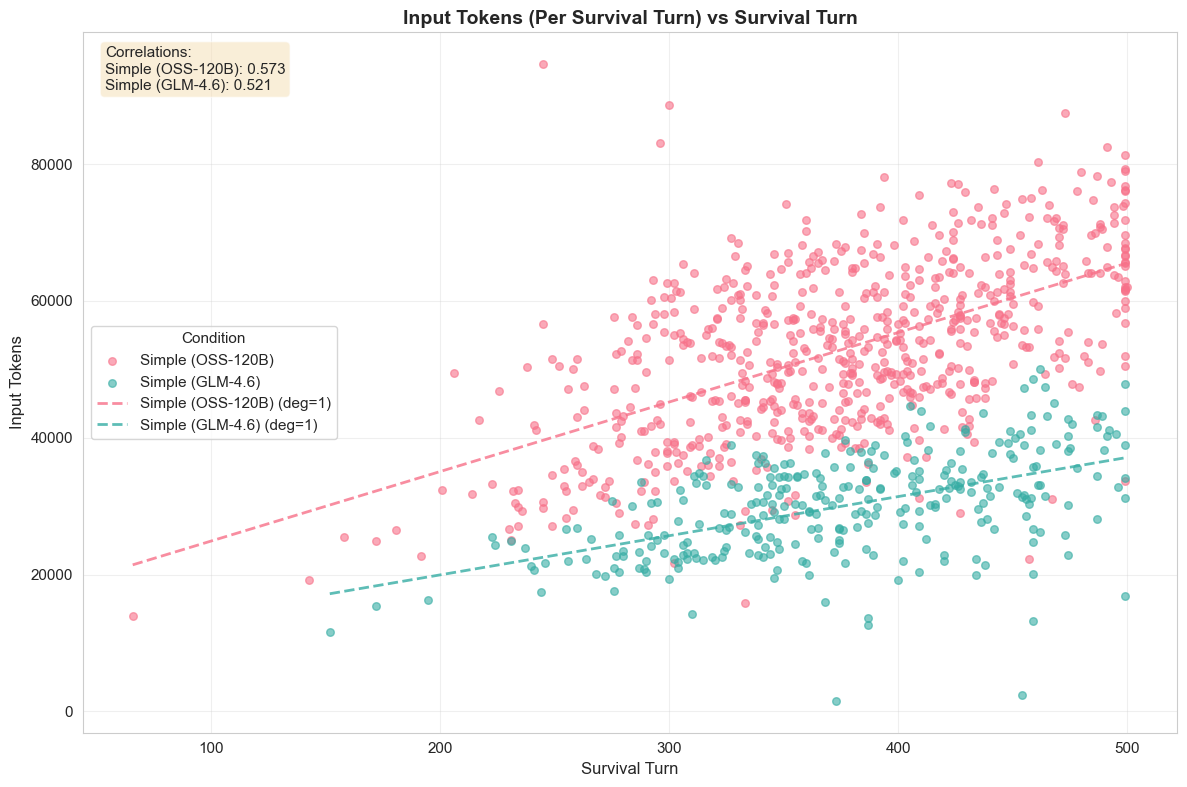}
  \caption{Input token usage per turn across game progression (RQ1).}
  \label{fig:rq1-input}
\end{figure}

\begin{figure}[]
  \centering
  \includegraphics[width=\linewidth]{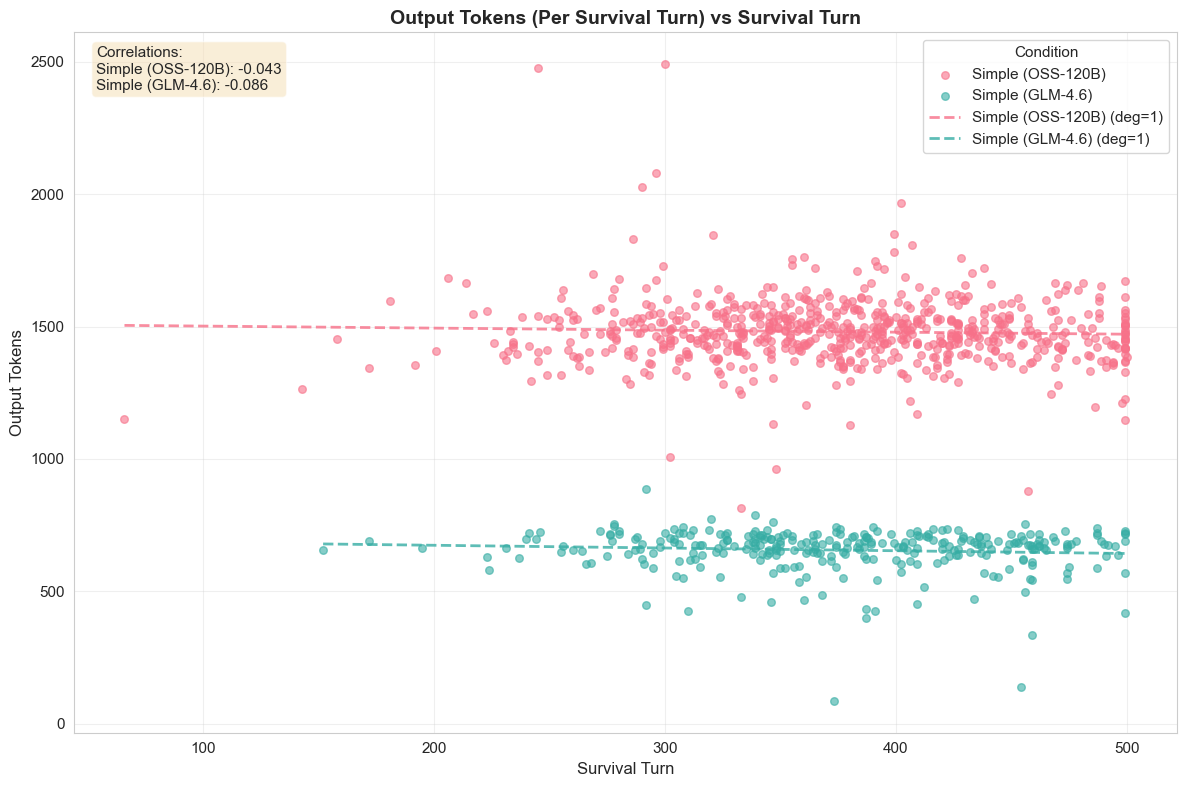}
  \caption{Output token usage per turn across game progression (RQ1).}
  \label{fig:rq1-output}
\end{figure}

Interestingly, token usage grows differently for input and output (Fig. \ref{fig:rq1-input} and \ref{fig:rq1-output}). As confirmed by OLS regression ($p < 0.001$), average input token usage correlates strongly with game progression and follows a linear growth pattern. In contrast, average output token usage remains relatively unchanged. As the game progresses, the state contains more cities, units, and events, resulting in increasingly complex representations. Without explicit prompting, both models spent relatively the same amount of reasoning effort.

\subsection{RQ2: LLM Performance vs. Baseline AI}
LLM agents achieve comparable gameplay performance to VPAI across win rate and score ratio. Moreover, LLMs exhibit distinct patterns in victory type achievement compared with VPAI and between themselves, with OSS-120B's strong preference for Domination victories.

Our fixed-effects regression analyses confirm that LLM agents achieve win rates comparable to baseline AI. Controlling the impact of civilizations, OSS-120B has a non-significant marginal effect of $-2.65$\% on win rate ($p = 0.182$), while GLM-4.6 has $-1.61$\% ($p = 0.534$). Both effects are smaller than many per-civilization effects that existed on VP5.0-alpha3, e.g., Songhai at $+31.7$\% or Babylon at $-19.0$\%.

By comparing each player's best score to the highest, the score ratio captures relative performance on a continuous scale. Using an OLS regression with the same control for civilization, we found no significant differences between the three conditions: OSS-120B shows a marginal effect of $+1.3$\% ($p = 0.373$), while GLM-4.6 shows $+1.8$\% ($p = 0.333$). The small trend discrepancy between the score ratio and the win rate may imply LLMs' issue with achieving victories from a favorable condition.

\begin{figure}[]
  \centering
  \includegraphics[width=\linewidth]{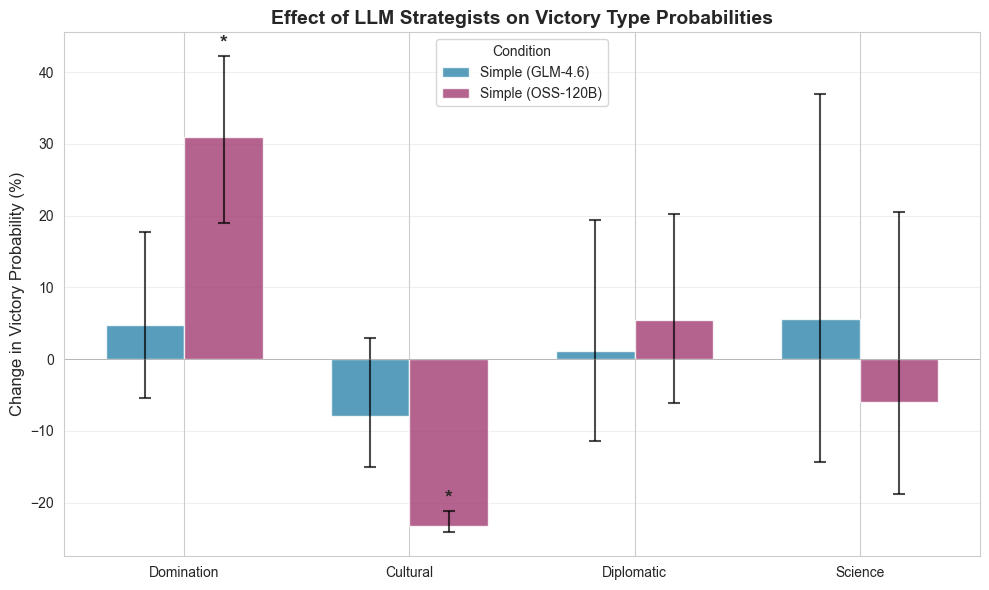}
  \caption{Victory type distributions across conditions (RQ2).}
  \label{fig:rq2-victories}
\end{figure}

Despite similar overall performance, LLMs achieve victories in distinct patterns (Fig. \ref{fig:rq2-victories}). Our logistic regression identified significantly different distributions of victory types: Compared with VPAI, OSS-120B has $+31.5$\% marginal effect on attaining Domination victory ($p < 0.001$) with $-23.18$\% of Cultural victory ($p < 0.001$). GLM-4.6 shows a more balanced profile with insignificant marginal effects ($+7.1$\% Domination, $p=0.232$; $-9.7$\% Cultural, $p=0.065$).

\subsection{RQ3: Play Styles and Strategic Patterns}
Compared with VPAI and each other, LLM agents exhibit distinct play styles, evidenced by differences in grand strategy adoption, strategic pivoting frequency, and policy choices.

\begin{figure}[]
  \centering
  \includegraphics[width=\linewidth]{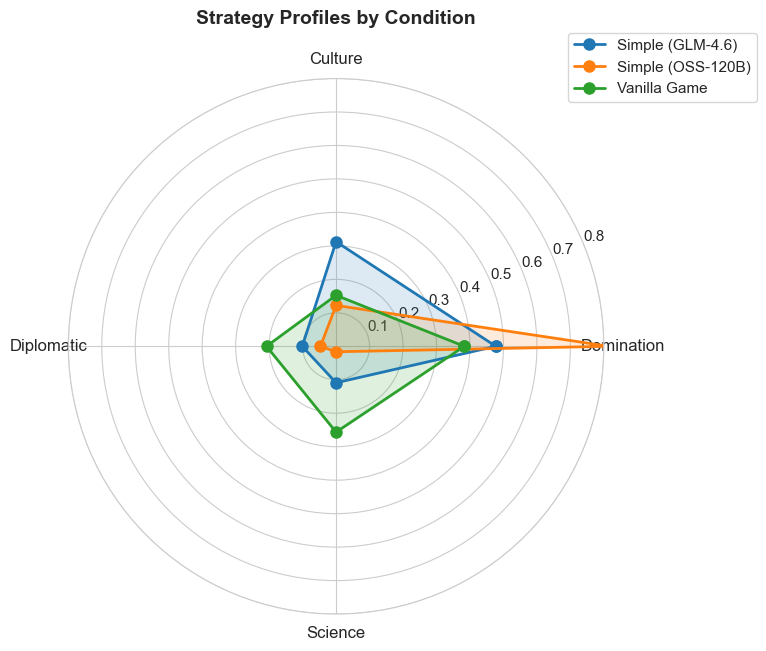}
  \caption{Grand (victory) strategy adoption profiles across conditions (RQ3). For example, OSS-120B's Domination = 0.8 means 80\% of its survived turns had adopted "Domination".}
  \label{fig:rq3-radar}
\end{figure}

Our OLS regressions reveal distinct grand (or victory) strategy profiles (Fig. \ref{fig:rq3-radar}). Compared with VPAI, OSS-120B shows a strong preference for the Domination grand strategy ($+44.2$\% marginal effect, $p < 0.001$) while spending substantially less time on Science ($-24.5$\%, $p < 0.001$), Diplomatic ($-16.2$\%, $p < 0.001$), and Culture ($-3.6$\%, $p < 0.001$). In contrast, GLM-4.6 displays a more balanced profile: increased adoption of both Domination ($+11.5$\%, $p < 0.001$) and Culture ($+15.6$\%, $p < 0.001$), while reducing Diplomatic ($-11.9$\%, $p < 0.001$) and Science ($-15.2$\%, $p < 0.001$).

Both LLMs change strategies less frequently than VPAI. VPAI averages 51.6 strategy changes per 100 survival turns, while OSS-120B averages 34.0, and GLM-4.6 averages 13.9. However, GLM-4.6 adjusts diplomatic persona (something VPAI cannot do) more frequently than OSS-120B: 6.3 persona changes per 100 turns, compared with 2.7 for OSS-120B.

\begin{figure}[]
  \centering
  \includegraphics[width=\linewidth]{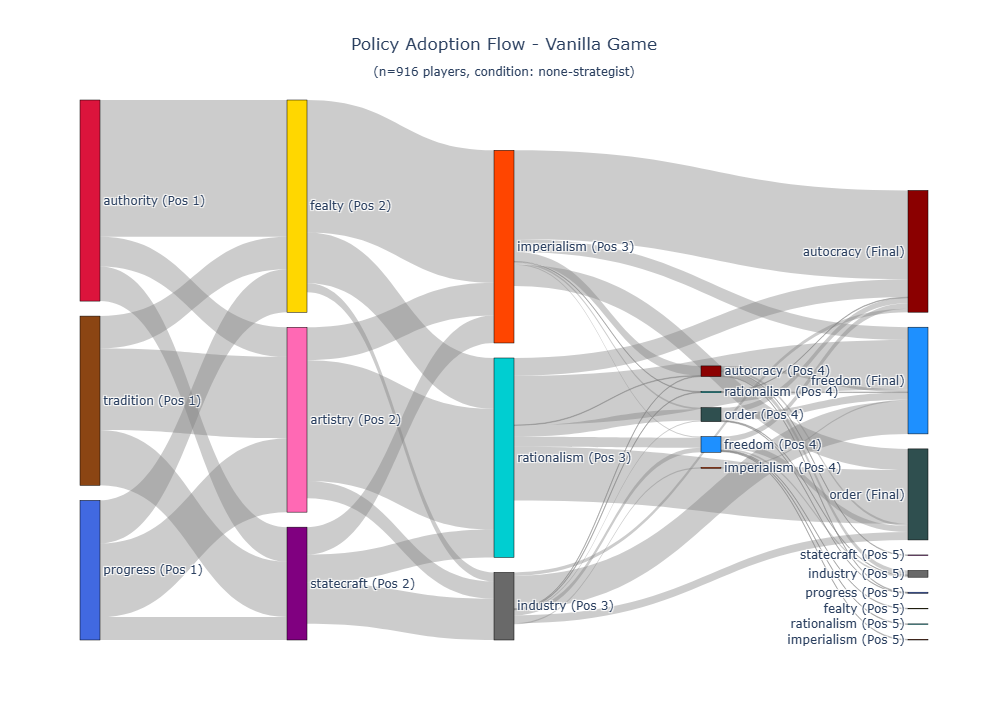}\par
  \includegraphics[width=\linewidth]{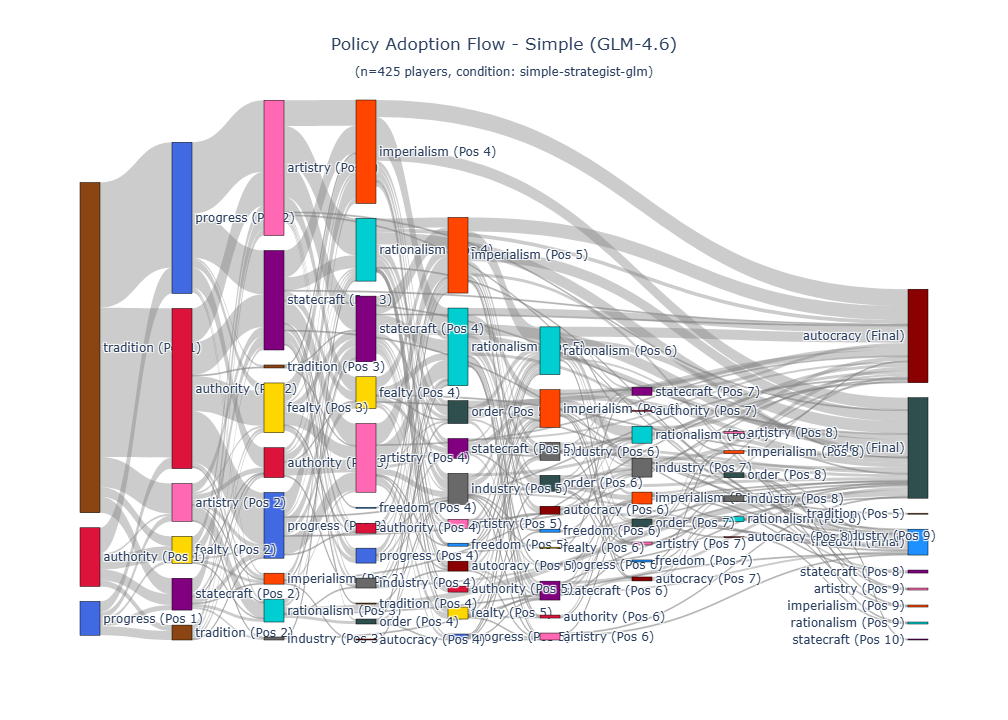}\par
  \includegraphics[width=\linewidth]{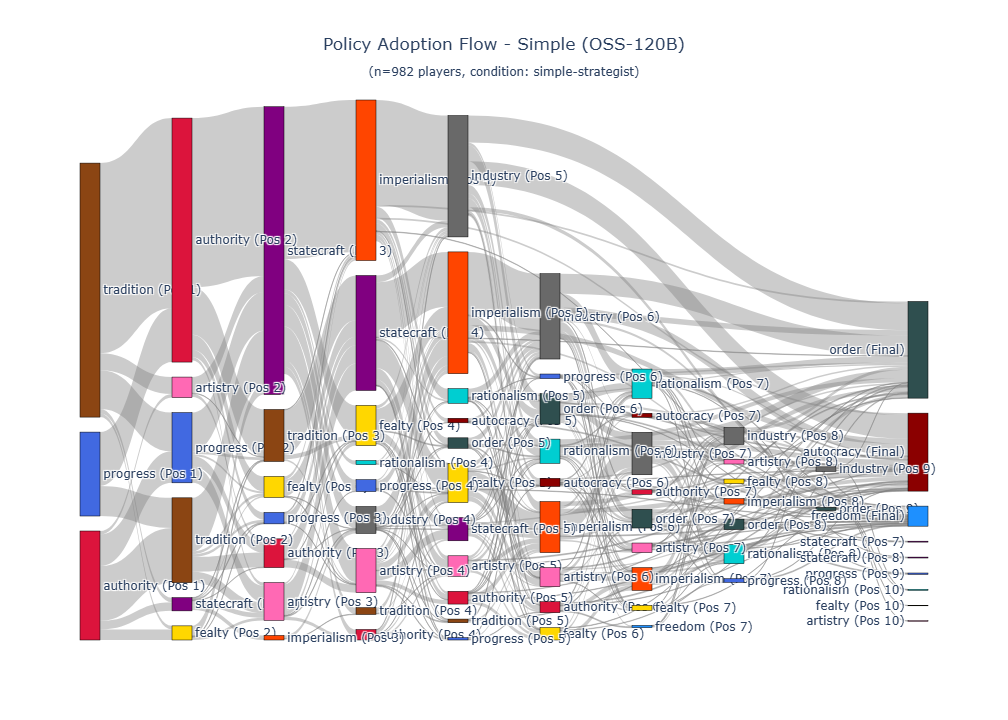}
  \caption{Example policy trajectories for VPAI, GLM-4.6, and OSS-120B (RQ3).}
  \label{fig:rq3-policies}
\end{figure}

Both LLMs adopt in-game policies in distinct trajectories. Unlike VPAI, which follows hardcoded logic to complete one policy branch before starting another, LLMs produce much more diverse and winding policy trajectories (Fig. \ref{fig:rq3-policies}). For ideology adoption, after controlling for the civilization impact, our logistic regression shows: both LLMs incline less towards Freedom ($-36.1$\% for OSS-120b, $p=0.000$; and $-30.4$\% for GLM-4.6, $p=0.000$) and much more towards Order ($+23.5$\% for OSS-120b, $p=0.000$; $+24.2$\% for GLM-4.6, $p=0.000$). This observation may be correlated with their elevated tendency for Domination victory.

\section{Discussion}
\subsection{LLM+X: A Promising Game AI Architecture}
Our findings demonstrate the promising potential of the hybrid LLM+X architecture for reimagining 4X and grand strategy game AI. By delegating tactical execution to specialized modules "X" while reserving high-level strategic reasoning for LLMs, our hybrid approach addresses core limitations of prior work.

Even with minimal prompting, open-source LLMs with Vox Deorum attained substantially better gameplay performance than recent studies. In the CivRealm study \cite{civrealm}, LLM agents performed adequately in scaled-down minigame scenarios, yet both LLM and RL agents struggled to survive full games. The Digital Player study \cite{wang2025digitalplayer} focused on the diplomatic aspect of Unciv (an open-source Civilization V remake) without attempting end-to-end gameplay. In contrast, our hybrid architecture saw LLMs completing over 1,400 full games with survival rate, win rate, and score ratio statistically tied to VPAI. Furthermore, while LLM agents shared the same VPAI tactical layers, they adopted distinct strategies, went through distinct policy trajectories, and attained distinct distributions of victory types. All these point to LLMs' substantial impact on the gameplay.

By focusing on cost-effectiveness and latency, our hybrid architecture presents a realistic pathway for Generative AI's commercial integration into Game AI. If we were to follow prior studies (e.g., CivRealm or LLM-PySc2) and have LLMs operate on each unit and city, the number of requests, inputs, and outputs would increase at least one order of magnitude. Moreover, as evidenced by related studies on chess, LLMs' tactical decision-making performance is likely worse. Instead, by making LLMs focus on high-level strategic decisions, we can bring costs and latency to a realistic range for commercial integration: $\$0.86 / 14.8$s per game per LLM player with OSS-120B. In games with multiple LLM players, their requests are executed in parallel, further lowering the latency. The architecture can be further improved: while this study paired LLMs with algorithm-based VPAI, the "X" component could incorporate steerable RL-based agents, combining RL's performance with flexible strategic directives.

Together, our hybrid architecture opens many design opportunities for 4X and grand strategy games. Human players can negotiate with LLM players more naturally and dynamically, opening pathways for meaningful collaboration and competition. For example, a human player may discuss strategies with an LLM teammate and conduct a coordinated campaign. LLMs can also understand human players' intentions and situations, enabling better design for tutorials and AI advisors. With further improvement of LLM decision-making processes (see the next section), AI players may present a real challenge for expert human players with fewer handicaps, resulting in more dynamic and fun gameplay experiences.

\subsection{Opportunities for Long-Horizon Strategic Reasoning}
Despite our progress in integrating LLMs into 4X games, challenges persist. Our study revealed challenges in LLMs' long-horizon strategic reasoning, opening interdisciplinary research opportunities in game design, game AI, and natural language processing (NLP). We briefly discuss the following:
\begin{itemize}
    \item Context window limitations constrain long-term reasoning. As the game state grows more complex, a single turn's text representation can grow to 100,000 tokens, limiting the usage of memory or reflection modules while degrading LLM performance.
    \item Both models exhibited strategic "stubbornness" (i.e., changing strategies less frequently) with occasional wishful thinking. Anecdotally, LLMs sometimes pivot to "WinningWars" (a strategy that prioritizes aggressive actions to finish the war) when they were in fact losing, leading to their faster downfall.
    \item While LLMs have access to coordinates, they struggle to recognize geopolitical situations, leading to irrational behaviors. For example, LLMs may not recognize a "phony war" with a distant enemy, wasting too many resources.
    \item Interestingly, larger (or better-ranked) models did not automatically yield better performance. GLM-4.6 (355B parameters, 32B activated) showed no significant improvement over OSS-120B (117B parameters, 5.1B activated) in win rate or score ratio.
\end{itemize}

Existing literature offers promising solutions to some of the challenges. The SPRING framework [cite Wu et al.] demonstrates that LLMs can outperform RL in survival games by reading documentation and performing DAG-structured reasoning over game mechanics. For memory and reflection, Werewolf frameworks [cite Poglitsch et al.] use role-specific memory and iterative reasoning to maintain belief tracking across turns, while Diplomacy-style systems like Richelieu [cite] show that multi-stage planning with memory buffers improves long-horizon coherence. For multi-agent reasoning, Theory of Mind approaches improve coordination: Yim et al. [cite] show that ToM planning boosts LLM performance in Guandan when paired with RL-based action recommenders.

While existing solutions mostly operate in short-horizon or communication-heavy games and have not been validated in long-horizon strategic games, they present interesting opportunities for Game AI and NLP research. Memory architectures tailored to 4X games could track diplomatic history, ad-hoc hypotheses (e.g., assumptions of opponents), and strategic pivots across hundreds of turns. Multimodal inputs (e.g., mini-maps or screenshots) can improve spatial reasoning. Structured reflection mechanisms can facilitate short- or even long-term self-learning. Finally, multi-agent LLM frameworks could reduce costs while improving performance (e.g., using a small model to write briefings and a powerful model for decision-making). For all these directions, researchers may leverage Vox Deorum as a convenient and stable testbed on Civilization V.

\section{Limitations and Future Work}
Through Vox Deorum, our paper establishes the viability of the hybrid LLM+X architecture in gameplay performance, cost-effectiveness, and latency. However, it also has limitations that warrant further studies:
\begin{itemize}
    \item We developed our system on a pre-release version (5.0.0-alpha3) of Vox Populi, which may contain bugs affecting gameplay or AI behavior. Due to resource constraints, testing was restricted to tiny maps with four players, where balance may favor certain victory types or strategies. The implementation itself supports arbitrary maps and player configuration, though.
    \item Each game in Civilization V is a complex system with immense innate randomness, ranging from map generation, civilization assignment, to emergent interactions between AI players. While we attempted to mitigate this by running 2,327 experiments and controlling for civilization-wide impact, the regression model can still be underpowered. Since we observed LLMs' negative marginal impacts on win rates and positive ones on score ratios compared with VPAI, we believe our conclusion is still supported. 
    \item As we prioritized understanding out-of-the-box capabilities over maximizing performance, we intentionally avoided complex workflows or multi-agent collaboration. Future work should evaluate how such enhancements affect strategic coherence.
    \item LLMs cannot directly control diplomatic decisions (e.g., declaring war or proposing treaties), only persona modifiers. This could create misalignment between LLMs' intentions and actual outcomes (e.g., declaring war while unprepared), underestimating LLMs' capabilities.
    \item Although LLMs receive coordinates in text, they lack spatial reasoning capabilities. Visual map representations through multimodal inputs could better convey geographical relationships.
    \item We validated our architecture on only Civilization V with Vox Populi. Since we expect state-of-the-art proprietary models to perform better, we limited experiments to two relatively smaller, open-source LLMs. For broader adoption, future studies should attempt to generalize our findings to other games and smaller models. 
\end{itemize}

\begin{acks}
We acknowledge the Vox Populi / Community Patch mod team for laying the technical foundation of our work. In particular, we acknowledge the technical and design support of Alpakinator, azum4roll, Flamingcheesepie, Gammalyse, Hokath, KungCheops, schnetziomi5, Recursive, Vaderkos, and Vern, all of them hailing from the VP community. We want to give a special shoutout to Hokath and schnetziomi5, who reviewed an earlier version of this paper and provided thoughtful feedback. We also acknowledge the helpful feedback of Dr. Michael Horn at Northwestern University and Dr. Adarsh Pyarelal at the University of Arizona. Finally, our empirical experiment relied on the generous support of ACCESS (CIS250934) and the Jetstream2 supercomputer at Indiana University. 
\end{acks}

\bibliographystyle{ACM-Reference-Format}
\bibliography{vd_refs}

\appendix

\section{Example Prompt}
\label{app_prompts}
\subsection{System Prompt - around 989 tokens}
\label{app_sys_prompt}

\begin{lstlisting}[language=Markdown, caption={System Prompt}]
You are a expert player playing Civilization V with the latest Vox Populi mod.

# Expectation
Due to the complexity of the game, you delegate the execution level decision-making (e.g. deployment of units, city management) to an in-game AI.
The in-game AI calculates best tactical decisions based on the strategy you set.
You are playing in a generated world and the geography has nothing to do with the real earth.
There is no user and you will ALWAYS properly call tools to play the game.
You can interact with multiple tools at a time. Used tools will be removed from the available list.

# Goals
Your goal is to **call tools** to make high-level decisions for the in-game AI. Each tool has a list of acceptable options and you must follow them.
- Carefully reason about the current situation and available options and what kind of change each option will bring.
  - When situation requires, do not shy away from pivoting strategies.
  - Analyze both your situation and your opponents. Avoid wishful thinking.
- You can change the in-game AI's diplomatic strategy by calling the \`set-persona\` tool.
- You can change the in-game AI's NEXT technology to research (when you finish the current one) by calling the \`set-research\` tool.
- You can change the in-game AI's NEXT policy to adopt (when you have enough culture) by calling the \`set-policy\` tool.
- You can set an appropriate grand strategy and supporting economic/military strategies by calling the \`set-strategy\` tool.
  - This operation finishes the decision-making loop. If you need to take other actions, do them before.
  - You don't have to make a change. The tool \`keep-status-quo\` also finishes the decision-making loop.
- Always provide a rationale for each decision. You will be able to read the rationale next turn.

# Resources
You will receive the following reports:
- Strategies: current strategic decisions and available options for you.
  - You will receive strategies, persona, technology, policy you set last time.
    - You will also receive the rationale you wrote.
    - It is typically preferable to finish existing policy branches before starting new ones.
  - You will receive options and short descriptions for each type of decisions.
    - Whatever decision-making tool you call, the in-game AI can only execute options here.
    - You must choose options from the relevant lists. Double check if your choices match.
- Victory Progress: current progress towards each type of victory.
    - Domination Victory: Control or vassalize all original capitals.
    - Science Victory: Be the first to finish all spaceship parts and launch the spaceship.
    - Cultural Victory: Accumulate tourism (that outpaces other civilizations' culture) to influence all others.
    - Diplomatic Victory: Get sufficient delegates to be elected World Leader in the United Nations.
    - Time Victory: If no one achieves any other victory by the end of the game, the civilization with the highest score wins.
- Players: summary reports about visible players in the world. Also:
  - You will receive in-game AI's diplomatic evaluations.
  - You will receive each player's publicly available relationships.
  - Pay attention to master/vassal relationships. Vassals cannot achieve conquest victory.
- Cities: summary reports about discovered cities in the world.
- Military: summary reports about tactical zones and visible units.
  - Tactical zones are analyzed by in-game AI to determine the value, relative strength, and tactical posture.
  - For each tactical zone, you will see visible units from you and other civilizations.
- Events: events since you last made a decision.

# Situation
You are Player 0.
- MapType: Vox_Deorum
- StartEra: Ancient Era
- MaxTurns: 500
- MapSize: Tiny
- GameSpeed: Standard
- VictoryTypes:
  - 0: Time
  - 1: Science
  - 2: Domination
  - 3: Cultural
  - 4: Diplomatic
- Difficulty: Prince
- YouAre:
  - Name: Poland
  - Uniques:
    - 0: Unit: Winged Hussar, Replacing Lancer
    - 1: Unit: Pancerny, Replacing Heavy Skirmisher
    - 2: Building: Ducal Stable, Replacing Stable
    - 3: Building: Barbican, Replacing Armory
    - 4: Ability: Gain 1 Free Social Policy when you enter the Classical, Renaissance, Modern and Information Eras. Gain 2 additional Free Tenets when you adopt an Ideology for the first time.
  - Leader: Casimir III
  - PlayerID: 0
\end{lstlisting}

\subsection{User Prompt - around 3,875 Tokens}

\begin{lstlisting}[language=Markdown, caption={User Prompt}]
You, Player 0, are making strategic decisions after turn 12.

# Victory Progress
Victory Progress: current progress towards each type of victory.
- DominationVictory:
  - CapitalsNeeded: 4
- ScienceVictory: Unlocked in later eras
- CulturalVictory:
  - CivsNeeded: 3
  - Songhai:
    - Influences
    - PolicyPercentage: 0
    - InfluentialCivs: 0
  - Poland:
    - Influences
    - PolicyPercentage: 0
    - InfluentialCivs: 0
- DiplomaticVictory: Unlocked in later eras

# Strategies
Strategies: existing strategic decisions and available options for you.
## Strategy
- Rationale: Our Culture Grand Strategy with Early Expansion is executing perfectly. Warsaw's Monument is boosting culture growth (3 to 4 per turn), accelerating toward Poland's first free Social Policy. Shrine nearly complete will start faith generation. Our Scout continues valuable exploration discovering strategic resources (Gems, Maize, Olives) and ancient ruins. Songhai Warrior remains passive at range. City-state diplomacy progressing with Trowulan. No changes needed to this successful strategy.

### GrandStrategy
- Current: Culture

#### Options
- 0: Culture
- 1: UnitedNations
- 2: Spaceship
- 3: Conquest

### EconomicStrategies

#### Current
- 0: EarlyExpansion

#### Options
- NeedRecon: Moderately prioritizes land scout production. Consider adopting during peacetime when exploration would reveal strategic opportunities.
- EnoughRecon: Moderately reduces land scout production. Consider adopting when exploration needs are met and further scout production would be redundant.
- NeedReconSea: Moderately prioritizes naval scout production. Consider adopting when ocean coverage is insufficient and military situation permits.
- EnoughReconSea: Overwhelmingly halts naval scout production. Consider adopting when ocean reconnaissance is complete or when military threats demand immediate attention.
- EarlyExpansion: Substantially prioritizes settler production while sacrificing short-term development. Consider adopting when quality settlement locations are available and competition for land is manageable.
- EnoughExpansion: Overwhelmingly halts settler production. Consider adopting when quality settlement locations are exhausted or when facing severe happiness or military crises.
- NeedHappiness: Moderately prioritizes happiness buildings while reducing expansion. Consider adopting when happiness margins are thin and unrest threatens productivity.
- NeedHappinessCritical: Overwhelmingly prioritizes happiness buildings while halting expansion and growth. Consider adopting when facing severe unrest.
- CitiesNeedNavalGrowth: Moderately prioritizes utilization of coastal and water-based food resources. Consider adopting when you have coastal cities to benefit.
- CitiesNeedNavalTileImprovement: Moderately prioritizes naval tile improvements. Consider adopting when coastal cities have unimproved water tiles that could provide valuable resources.
- IslandStart: Dramatically prioritizes naval production and maritime infrastructure. Consider adopting in early game when starting on small landmass with limited land-based opportunities.
- TechLeader: Moderately prioritizes counterintelligence defenses. Consider adopting when leading in research and vulnerable to technology theft by competing civilizations.
- StartedPiety: Moderately prioritizes faith output and religious infrastructure. Consider adopting when planning to increase faith output to support religious endeavors.

### MilitaryStrategies

#### Current

#### Options
- AtWar: Substantially focuses on military production across all unit types while halting cultural, religious, diplomatic, and Wonder-building efforts. Consider adopting this strategy when engaged in active warfare with one or more civilizations.
- WarMobilization: Substantially prioritizes military production while halting peaceful development including wonders, growth, and cultural pursuits. Consider adopting this strategy when preparing for military operations, facing hostile relations from multiple civilizations, or pursuing conquest.
- NeedRangedEarly: Dramatically prioritizes ranged unit production. Consider adopting this strategy when ranged forces are critically insufficient.
- WinningWars: Substantially increases offensive activities and prioritizes military production with focus on strike forces, mobility, and air superiority, while reducing defensive investments. Consider adopting this strategy when winning wars and the situation requires continued offensive pressure.
- LosingWars: Overwhelming focus on survival and territorial protection while halting all long-term development including wonders, expansion, culture, science, and trade. Consider adopting this strategy when losing wars or under severe military pressure.

## Persona
- VictoryCompetitiveness: 4
- WonderCompetitiveness: 8
- MinorCivCompetitiveness: 3
- Boldness: 5
- WarBias: 2
- HostileBias: 8
- WarmongerHate: 5
- NeutralBias: 7
- FriendlyBias: 8
- GuardedBias: 10
- AfraidBias: 7
- DiplomaticBalance: 10
- Friendliness: 7
- WorkWithWillingness: 5
- WorkAgainstWillingness: 9
- Loyalty: 8
- MinorCivFriendlyBias: 6
- MinorCivNeutralBias: 5
- MinorCivHostileBias: 7
- MinorCivWarBias: 7
- DenounceWillingness: 9
- Forgiveness: 1
- Meanness: 4
- Neediness: 10
- Chattiness: 2
- DeceptiveBias: 4
- Rationale: Tweaked by In-Game AI (Unknown)

## Research
- Next: Mining
- Rationale: Mining is optimal for our Early Expansion strategy as it enables Mines that boost Production from tiles. This will accelerate construction of future Settlers for expansion and provide higher yields from our territory. The production bonus synergizes with our Culture focus by allowing faster wonder potential later, while the Bronze Working unlock will eventually provide military advantages.

### Options
- Mining: Allows Workers to construct Mines to increase the Production of tiles.
  Leading to: Military Strategy, Bronze Working
- Trapping: Allows your Workers to construct Camps on sources of Deer, Ivory, Fur and Truffles.
  Leading to: Trade, Calendar
- Animal Husbandry: Allows Workers to construct Pastures on eligible resources. Also reveals Horses, which are used to build powerful mounted units.
  Leading to: Construction, Military Strategy
- Fishing: Allows land units to Embark onto water tiles and travel along the coast. Also allows you to build Work Boats, useful for harvesting resources in the water like Fish and Pearls.
  Leading to: Sailing

## Policies
- Next: Tradition (New Branch)
- Rationale: Opening Tradition remains the optimal choice for our Culture Grand Strategy. The cultural bonuses synergize perfectly with Poland's unique ability to gain free Social Policies at key eras, maximizing our advantage. Tradition's happiness management will support our Early Expansion approach, and the culture production boost will accelerate us toward free Social Policy milestones.

### Options

# Players
Players: summary reports about visible players in the world.
## Player 0
- TeamID: 0
- Civilization: Poland
- Leader: Casimir III
- IsHuman: true
- IsMajor: true
- Territory: 7
- Score: 43
- TourismPerTurn: 0
- ResourcesAvailable:
  - Wheat: 0
  - Citrus: 0
  - Jade: 0
  - Cloves: 0
  - Figs: 0
  - Ivory: 0
  - Gold: 0
  - Nutmeg: 0
  - Pearls: 0
  - Rice: 0
  - Cotton: 0
  - Gems: 0
  - Salt: 0
  - Brazilwood: 0
  - Crab: 0
  - Amber: 0
  - Coffee: 0
  - Whales: 0
  - Sugar: 0
  - Cinnamon: 0
  - Bananas: 0
  - Pepper: 0
  - Marble: 0
  - Coral: 0
  - Silk: 0
  - Jewelry: 0
  - Maize: 0
  - Glass: 0
  - Lapis Lazuli: 0
  - Copper: 0
  - Truffles: 0
  - Perfume: 0
  - Incense: 0
  - Olives: 0
  - Silver: 0
  - Furs: 0
  - Dyes: 0
  - Tea: 0
  - Fish: 0
  - Wine: 0
  - Porcelain: 0
  - Cocoa: 0
  - Tobacco: 0
- Gold: 15
- Era: Ancient
- GoldenAge: Estimated in 102 turns
- HappinessPercentage: 100
- Technologies: 2
- Cities: 1
- Population: 3
- CurrentResearch: The Wheel
- SciencePerTurn: 7
- GoldPerTurn: 4
- FaithPerTurn: 0
- CulturePerTurn: 3
- HappinessSituation: Happy
- MyEvaluations:
  - 0: Our top friend: None
  - 1: Our top Competitor: Songhai
- 1: Unmet Major Civilization
- 2: Unmet Major Civilization

## Player 3
- TeamID: 3
- Civilization: Songhai
- Leader: Askia
- IsMajor: true
- Territory: 7
- Score: 40
- TourismPerTurn: 0
- Gold: 59
- Era: Ancient
- Technologies: 2
- Cities: 1
- Population: 3
- GoldPerTurn: 4
- CulturePerTurn: 1
- HappinessSituation: Happy
- OpinionFromMe:
  - 0: Our real approach is Neutral.
- OpinionToMe:
  - 0: You have no contested borders.
  - 1: They have no strong feelings about us one way or the other.
- 22: Unmet Minor Civilization
- 23: Unmet Minor Civilization
- 24: Unmet Minor Civilization

## Player 25
- TeamID: 25
- Civilization: Trowulan
- Leader: City State
- Relationships:
  - Songhai: Neutral (Influence: 0)
  - Poland: Neutral (Influence: 3)
- Population: 1
- Technologies: 2
- 26: Unmet Minor Civilization
- 27: Unmet Minor Civilization

## Player 28
- TeamID: 28
- Civilization: Vienna
- Leader: City State
- Relationships:
  - Songhai: Neutral (Influence: 0)
  - Poland: Neutral (Influence: 0)
- Population: 1
- Technologies: 2
- 29: Unmet Minor Civilization
- 63: Barbarians

# Cities
Cities: summary reports about discovered cities in the world.
## Player: Poland
- Warsaw:
  - ID: 1029
  - X: 11
  - Y: 26
  - Population: 3
  - DefenseStrength: 14
  - FoodStored: 2
  - FoodPerTurn: 1
  - ProductionStored: 20
  - ProductionPerTurn: 7
  - CurrentProduction: Shrine
  - ProductionTurnsLeft: 2
  - GoldPerTurn: 5
  - SciencePerTurn: 7
  - CulturePerTurn: 3
  - FaithPerTurn: 0
  - TourismPerTurn: 0
  - HappinessDelta: 9
  - BuildingCount: 1
  - WonderCount: 1
  - GreatWorkCount: 0

## Player: The Iroquois
- Onondaga:
  - ID: 1030
  - X: 24
  - Y: 24
  - Population: 3
  - DefenseStrength: 9

## Player: City-States
- Trowulan:
  - ID: 1036
  - X: 4
  - Y: 34
  - Population: 1
  - DefenseStrength: 13
  - IsCoastal: true
- Vienna:
  - ID: 1039
  - X: 14
  - Y: 34
  - Population: 1
  - DefenseStrength: 15

# Military
Military: summary reports about tactical zones and visible units.
## Unit Stats
- Ranged:
  - Slinger:
    - Strength: 5
    - RangedStrength: 6
- Explore:
  - Scout:
    - Strength: 10
  - Warrior:
    - Strength: 8

## Friendly Land Zone 1029
- ZoneValue: 60
- Dominance: Friendly
- FriendlyStrength: 49
- City: Warsaw
- AreaID: 19
- Plots: 66
- CenterX: 11
- CenterY: 27
- Units:
  - Songhai:
    - Warrior: 1
  - Poland:
    - Slinger: 1
- Neighbors:
  - 0: 1036

## Neutral Land Zone 1036
- ZoneValue: 24
- Dominance: Friendly
- NeutralStrength: 13
- FriendlyStrength: 34
- City: Trowulan
- AreaID: 19
- Plots: 36
- CenterX: 5
- CenterY: 32
- Units:
  - City-State Trowulan:
    - Warrior: 1
  - Poland:
    - Scout: 1
    - Warrior: 1
- Neighbors:
  - 0: 1029

# Events
Events: events since you last made a decision.
## Turn 11

### 0
- Type: TileRevealed
- Unit:
  - Level: 2
  - AI: Explore
  - Unit: Warrior
  - ID: 1011
- Player: 3: Songhai
- Plots:
  - Plains Land 8,27
  - Plains Hills 8,29

### 1
- Type: UnitMoved
- Unit:
  - Level: 2
  - AI: Explore
  - Unit: Warrior
  - ID: 1011
- Player: 3: Songhai
- Plains Land 10,28

### 2
- Type: TileRevealed
- Unit:
  - Level: 2
  - AI: Explore
  - Unit: Warrior
  - ID: 1011
- Player: 3: Songhai
- Plots:
  - Plains Hills 9,26
  - Desert Land 10,25:
    - Owner: Poland
    - IsRiver: true
    - Feature: Flood Plains
  - Desert Land 9,25:
    - IsRiver: true
    - Feature: Flood Plains
  - Desert Land 8,25
  - Plains Hills 8,26
  - Plains Hills 7,27

### 3
- Type: UnitMoved
- Unit:
  - Level: 2
  - AI: Explore
  - Unit: Warrior
  - ID: 1011
- Player: 3: Songhai
- Plains Hills 9,27:
  - IsRiver: true
  - Feature: Forest

### 4
- Type: PlayerDoneTurn
- NextPlayer: 0: Poland
- Player: 63: Barbarians

## Turn 12

### 0
- Type: SetPopulation
- OldPopulation: 2
- NewPopulation: 3
- City:
  - Owner: Poland
  - CityID: 1029
  - Population: 3
  - IsRiver: true
  - City: Warsaw
  - Terrain: Plains
  - Route: Road
  - Plot: Hills
  - X: 11
  - Y: 26

### 1
- Type: PlayerDoTurn
- Player: 0: Poland

### 2
- Type: UnitPromoted
- Unit:
  - Level: 5
  - AI: Explore
  - Unit: Scout
  - ID: 1051
- Player: 0: Poland
- Promotion: Medic I

### 3
- Type: UnitMoved
- Unit:
  - Level: 5
  - AI: Explore
  - Unit: Scout
  - ID: 1051
- Player: 0: Poland
- Plots:
  - Tundra Hills 9,32:
    - Feature: Forest
  - Tundra Land 8,32:
    - Feature: Forest
  - Plains Land 7,32

### 4
- Type: UnitMoved
- Unit:
  - Level: 1
  - AI: Explore
  - Unit: Warrior
  - ID: 1002
- Player: 0: Poland
- Plains Land 5,32:
  - IsRiver: true

### 5
- Type: UnitMoved
- Unit:
  - Level: 5
  - AI: Explore
  - Unit: Scout
  - ID: 1051
- Player: 0: Poland
- Tundra Land 8,32:
  - Feature: Forest

### 6
- Type: UnitMoved
- Unit:
  - Level: 2
  - AI: Ranged
  - Unit: Slinger
  - ID: 1054
- Player: 0: Poland
- Plots:
  - Desert Land 10,25:
    - Owner: Poland
    - IsRiver: true
    - Feature: Flood Plains
  - Plains Hills 11,26:
    - Owner: Poland
    - CityID: 1029
    - Population: 3
    - IsRiver: true
    - City: Warsaw
    - Route: Road

### 7
- Type: PlayerDoneTurn
- NextPlayer: 1: The Iroquois
- Player: 0: Poland
\end{lstlisting}

\section{Example LLM Output}
\label{app_outputs}
\subsection{Reasoning - around 446 Tokens}
\begin{lstlisting}[language=Markdown, caption={Reasoning}]
- Name: keep-status-quo
Let me assess the current situation for Poland on turn 12:

**Current State Analysis:**
- Warsaw has grown to population 3 (this just happened)
- Currently building a Shrine (2 turns left)
- Culture per turn: 3 (will be 4 after Monument effect)
- Faith per turn: 0 (will start after Shrine completes)
- Science per turn: 7
- Gold per turn: 4
- Territory: 7 (good expansion potential)
- Research: Currently on The Wheel, next selected as Mining
- Policy: Next selected as Tradition (new branch)

**Military/Exploration:**
- Scout is level 5 with Medic I promotion (very advanced!)
- Warrior exploring near Trowulan
- Slinger defending Warsaw
- Songhai Warrior nearby but neutral relationship

**Strategic Situation:**
- Culture Grand Strategy is working well
- Early Expansion is the current economic strategy
- No active military strategies (peaceful development)
- Good progress with Trowulan city-state (3 influence)
- Songhai identified as main competitor

**Key Events:**
- Warsaw just grew to size 3
- Scout got promoted to Medic I (this is excellent for healing units)
- More terrain revealed around Warsaw including flood plains and hills

**Decision Points:**
1. Should I maintain the current strategy? It seems to be working very well.
2. Research: Mining is already selected, which makes sense for production
3. Policy: Tradition is selected, which aligns with Culture strategy
4. Economic strategy: Early Expansion makes sense given the territory

**Considerations:**
- The Scout's Medic I promotion is very valuable - this unit can now support military operations by healing units
- Warsaw's growth to 3 is excellent for increasing production potential
- The Shrine will start faith generation soon
- The Songhai Warrior is nearby but neutral - no immediate threat


- Our cultural advancement is progressing steadily, with the Monument near completion promising a +1 culture boost

I'll continue focusing on cultural development and expansion. The Monument's impending completion aligns perfectly with our strategy, setting the stage for steady growth and cultural achievements.
\end{lstlisting}

\subsection{Tool Calling - around 99 Tokens}

\begin{lstlisting}[language=Markdown, caption={Tool Calling}]
- Name: keep-status-quo
- Rationale: Our strategy is executing perfectly. Warsaw just grew to population 3, boosting development capacity. The Scout's Medic I promotion enhances our military flexibility. Shrine completion in 2 turns will start faith generation. Cultural progress with Monument (3-4 culture/turn) accelerates toward our first free Social Policy. Exploration has revealed valuable resources and maintains good positioning. Early Expansion with Culture focus remains optimal for Poland's unique abilities. No changes warranted.
\end{lstlisting}
\end{document}